\documentclass[journal]{IEEEtran}

\usepackage{booktabs}

\ifCLASSINFOpdf
\usepackage[pdftex]{graphicx}
\usepackage{xcolor}
\usepackage{booktabs}

\else
\fi

\usepackage{adjustbox}
\usepackage{tikz}
\usetikzlibrary{arrows.meta, positioning, shapes.geometric, calc}

\begin{document}
%


\title{Modular Anthropomorphic Hand Design via Multi-Parameter Finger Benchmarking and Selection}


%
%

\author{Yu~Zhang{*},
        Huijiang~Wang{*}
        and~Josie~Hughes,~\IEEEmembership{Member,~IEEE}
        \vspace{-2em}
\thanks{The CREATE Lab, Institute of Mechanical Engineering, Swiss Federal Institute of Technology in Lausanne (EPFL), CH-1015, Lausanne, Switzerland. (e-mails: yu.zhang@epfl.ch; huijiang.wang@epfl.ch; josie.hughes@epfl.ch).}
\thanks{*Equal Contribution. \ Correspondence: Huijiang Wang.}}

\maketitle

\begin{abstract}
Designing anthropomorphic dexterous robotic hands remains challenging as the design space straddles morphology, actuation, and sensing properties, and performance metrics span both task-dependent and task-agnostic. Existing optimization methods are often unstructured or consider only a single performance metric, limiting systematic comparison and targeted refinement.
While the design considerations of the entire hand are significant, the individual finger properties play a key role in dexterity. By developing a robotic hand platform where fingers can be modularly integrated into a full teleoperated hand, we propose that optimizing the fingers can significantly improve overall hand performance.
This approach enables rapid screening of different finger-level prototypes through a number of quantitative benchmarks before their integration into the hand for task-level validation. Candidate finger designs (incorporating variations in joint, bone, skin, and sensor placement) are assessed using both mechanism-oriented and task-relevant metrics, which establish a quantitative link between component design and full hand embodiment.
The framework is validated through the development of an anthropomorphic robotic hand with optimized fingers, demonstrating how these fingers enable performance improvements across tasks, including multi-object grasping and light bulb screwing.
\end{abstract}

\begin{IEEEkeywords}
Anthropomorphic hand, modular design, robotic teleoperation, dexterous manipulation.\vspace{-1em}
\end{IEEEkeywords}

%
\IEEEpeerreviewmaketitle

\section{Introduction}

Dexterous robotic manipulation is important to enable meaningful task completion by robots operating in unstructured environments. Compared with parallel grippers and task-specific end effectors, anthropomorphic hands are of particular interest as they support a broader set of grasping and in-hand interaction modes, including enveloping grasping, in-hand reorientation and multi-contact stabilization \cite{bullock2012hand,liconti2026benchmark}. These capabilities become critical when robots move beyond repetitive pick and place and must handle diverse objects under uncertain contact conditions with greater adaptability and finer interaction control \cite{11455323,huang2025human}. Whilst control methods and AI can advance manipulation capabilities, the design of robotic hands still dominates manipulation capability, adaptability and task coverage.

However, designing dexterous robotic hands remains fundamentally difficult due to the considerable design space that includes morphology, joint mechanics, skin properties, and sensor placement.
Secondly, evaluating different hand designs requires multiple performance metrics, including workspace, force output, conformability, and other factors that collectively determine the dexterous capabilities of the hand~\cite{puhlmann2022rbo,jiang2025critical,fabisch2025robots}.
A single design choice often affects multiple performance dimensions at once, leading to trade-offs. For example, joint architecture and tendon routing influence both reachable workspace and force transmission \cite{kim2021integrated}, and surface material affects both grasp frictional conditions and contact compliance \cite{linghu2025versatile,lyu2025humanoid}. Finger-level performance transfers to hand-level dexterity, and whole-hand performance emerges from the interaction among fingers, palm structure, actuation strategy and task context \cite{teeple2021active,wang2024human,lu2025folding}. Therefore, dexterous hand design is not simply a matter of assembling components, but a high-dimensional and tightly coupled system design problem.

\begin{figure}[t]
    \centering
    \includegraphics[width=1\linewidth]{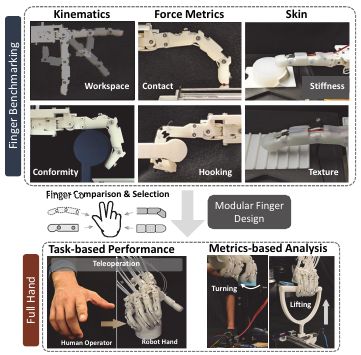}
    \caption{Overview of the iterative, modular design framework from individual fingers to the complete hand system.}
    \vspace{-1.2em}
    \label{fig:1_overview}
\end{figure}

\begin{figure*}[!h]
    \centering
    \includegraphics[width=0.8\linewidth]{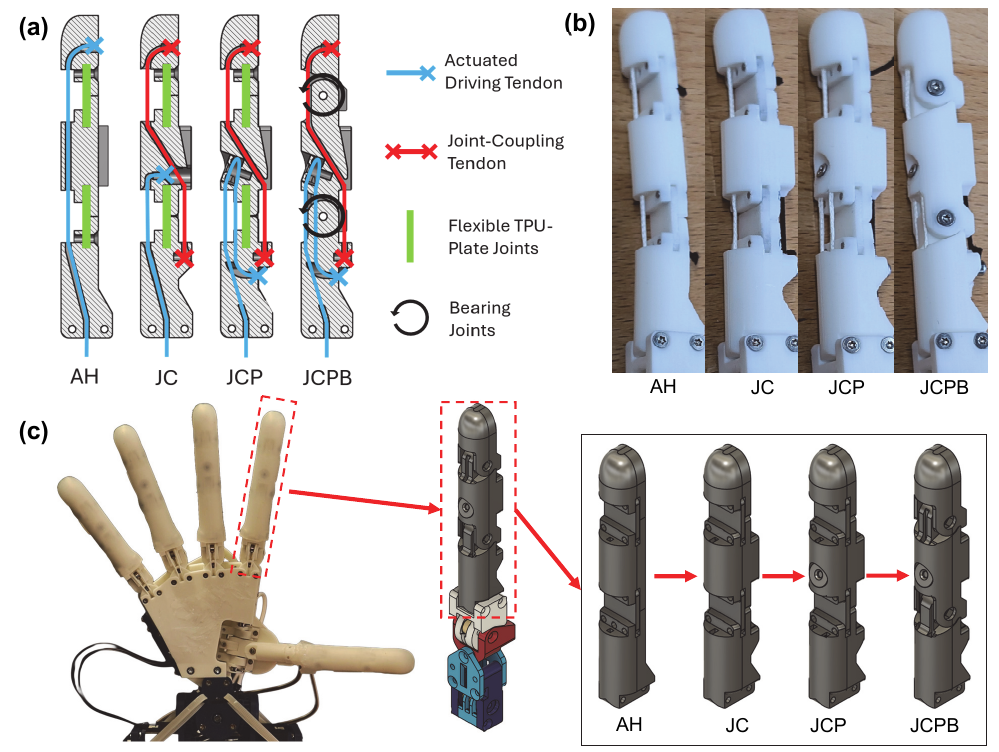}
    \caption{Finger modules and modular hand platform. (a) Schematics of the four finger designs. (b) Fabricated finger modules. (c) Modular hand assembled with interchangeable finger modules.}
    \vspace{-1.2em}
    \label{fig:2_finger}
\end{figure*}

Despite significant progress in the physical capabilities of robotic hands, many of these have arisen from loosely structured design processes. In a common workflow, a hand is designed based on prior experience, evaluated through a limited set of demonstrations, and then revised locally after failure cases are observed. While iteration does occur in practice, it is often implicit, weakly standardized, and difficult to reproduce across design variants \cite{xu2013low}. Simulation-based design can make this process more systematic by screening candidate geometries before fabrication. Existing studies have used finite-element analysis, topology optimization, and morphology--control co-design to improve soft robotic fingers, grippers, and hands~\cite{elsayed2014finite,chen2018topology,deimel2017automated,liu2020optimal}. These approaches show that geometry, compliance, and control parameters can be explored before hardware realization, reducing the reliance on purely trial-and-error prototyping.
However, accurate simulation remains difficult for tendon-driven anthropomorphic hands with compliant skins, where nonlinear contact, tendon-joint compliance, and skin-sensor coupling can substantially affect real performance~\cite{mannam2023designing}. 

An alternative approach is real-world evaluation and benchmarks. These include task-agnostic metrics, such as workspace, Kapandji score, force output, slip resistance, and tactile sensitivity, which provide controlled measures of low-level capability~\cite{chalon2010thumb,falco2020benchmarking}. Task-focused benchmarks, such as YCB protocols \cite{calli2015benchmarking}, NIST assembly task boards~\cite{kimble2020benchmarking}, and recent dexterity benchmarks~\cite{liconti2026benchmark,cruciani2020benchmarking}, assess whether a complete hand system can perform representative manipulation tasks. Whilst these benchmarks are useful for evaluation, these two levels remain weakly connected because low-level metrics do not directly explain task success, while task-level scores often mix the effects of hand morphology, control and sensing, and operator skill. Therefore, existing benchmarks offer limited guidance on how finger- or skin-level design choices should be selected to improve full-hand performance.

A further challenge in robotic hand development is ease of iterating across hardware.  
At the hardware level, many robotic hands are developed as full integrated systems which are physically coupled and interlinked \cite{park2020open}. This makes it difficult to isolate the effects of design variables, replace modules, perform controlled comparisons, or efficiently refine specific subsystems \cite{ma2013modular,xu2013low}. 

In this work, we address these limitations through an iterative, modular framework and hardware setup for dexterous robotic hand design. We hypothesize that optimizing individual finger modules within a reconfigurable hand platform directly translates to enhanced full-hand performance. This approach enables more targeted refinement of fingers across a wider design space and a broader range of performance metrics than would be possible through iterating on the entire hand assembly.
Following the framework in Fig. \ref{fig:1_overview}, we benchmark and select modular fingers and their corresponding skins through a quantitative scoring system to evaluate candidate designs through both mechanism-oriented and task-relevant metrics. The former captures kinematic workspace, force-related behavior, stiffness, and contact response. The latter captures contact conformity and various force metrics.  
These results guide finger selection across skin and mechanism, before they can be rapidly integrated into a full hand.
Integration of fingers is enabled through a palm unit which allows integration and connection of finger modules, enabling rapid exchange of fingers.
The full hand can then be analyzed through quantitative teleoperated task-based performance metrics, which allow the low-level finger improvements to assessed with respect to improvements in full hand dexterity and force application.
Through this design approach, the optimized JCPB hand achieved an \(83.3\%\) higher lateral lifting capacity, an approximately \(32\%\) higher full-hand grasping force, and an \(87.5\%\) larger knob-turning range compared with our baseline.

\section{Modular Hand Platform}
\label{sec2:MaterialMethod}

A modular hand platform was developed for systematic evaluation of finger structures and soft skin designs (Fig.~\ref{fig:2_finger}). Based on the tendon-driven ADAPT hand~\cite{junge2025spatially}, this retains many elements of the design (the palm, actuation unit, Bowden tube routing, and metacarpophalangeal(MCP) joint design) but incorporates an interface that allows for attachment to interchangeable fingers. Each modular finger then exists of the proximal interphalangeal (PIP) and distal interphalangeal (DIP) joints. The finger module can then be fixed by four screws to the hand and tendons connected. 
This allows the finger design to be varied, tested separately, and then integrated into the full anthropomorphic hand for testing via teleoperation or direct position control. 



\subsection{Finger Structural Designs}
Starting from our ADAPT hand (AH) baseline finger, we developed a further four bone and joint designs and eight swappable skins. 
These four designs reflect different degrees of friction reduction and under-actuation to explore various trade-offs in joint design and tendon routings. The four fingers share the same motor box and Bowden tube actuation mechanism~\cite{junge2025spatially}, which allows direct comparison between the previous ADAPT hand and the proposed hand equipped with the optimized finger and skin design. 

The four base finger designs include the ADAPT hand (AH) baseline finger, the joint coupled (JC) finger, the joint coupled finger with an integrated pulley (JCP), and the joint coupled finger with an integrated pulley and bearing joints (JCPB), shown in Fig.~\ref{fig:2_finger}. All designs used a single active flexion tendon and shared the same nominal PIP and DIP joint locations, phalanx lengths, and finger width. 
The rigid phalanges' morphology was fixed, enabling the comparison to focus on tendon routing, joint coupling, and joint implementation.
Whilst we explore these four designs, our approach could be used across a wider range of designs. 
The four finger designs are now briefly introduced:

\subsubsection{ADAPT Hand Baseline Finger}

The AH baseline finger follows the finger design of the ADAPT Hand V2~\cite{junge2025spatially}. A single tendon underactuates the PIP and DIP joints, which are constructed from flexible thermoplastic polyurethane (TPU) elements. This design provides passive adaptability, but the PIP and DIP joint motions cannot be specified independently.

\subsubsection{Joint Coupled Finger}

The JC finger introduces a passive tendon coupling between the PIP and DIP joints. The primary tendon actuates the PIP joint, while DIP flexion is generated through a secondary tendon routed across the PIP joint and attached to the distal phalanx. This coupling reduces the effective distal degrees of freedom and enables more defined fingertip motion.

\subsubsection{Joint Coupled Finger with Integrated Pulley}

The JCP finger preserves the PIP and DIP coupling mechanism and adds a pulley on the middle phalanx. This routing increases the actuation force of the PIP joint while maintaining coupled motion.

\subsubsection{Joint Coupled Finger with Integrated Pulley and Bearing Joints}

The JCPB finger uses the same tendon routing as the JCP finger design, but replaces the flexible TPU joints with miniature ball bearing joints. This modification reduces elastic resistance, viscoelastic damping, and friction-related loss in the joints and improves passive extension of the PIP and DIP joints driven by elastic strings.

\begin{figure}[t]
    \centering
    \includegraphics[width=1\linewidth]{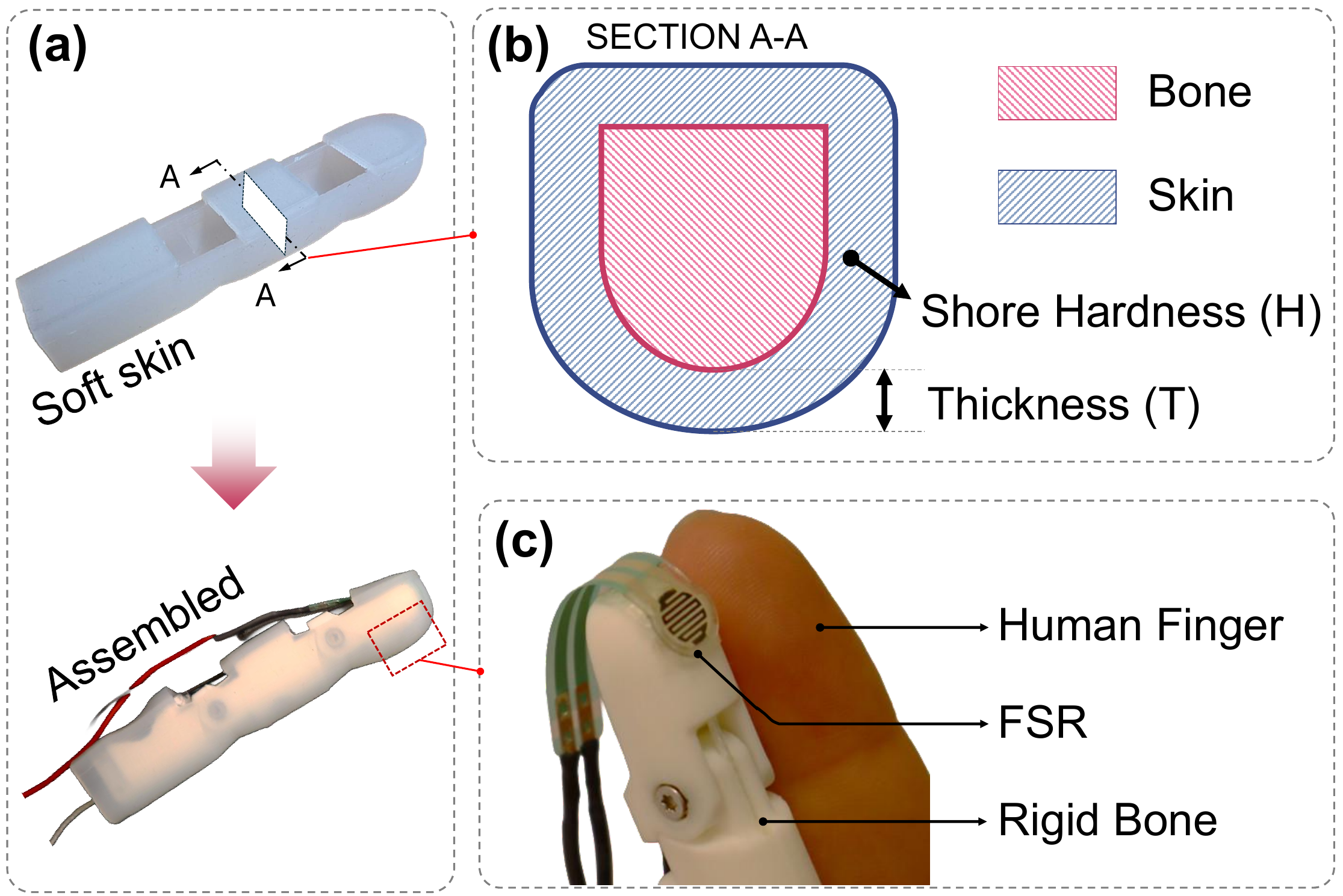}
    \caption{Soft skin module and sensing setup. (a) Modular soft skin before and after installation on the finger. (b) Cross-sectional schematic showing the rigid bone, soft skin, and the skin design variables, shore hardness \(H\) and thickness \(T\). (c) A force-sensitive resistor (FSR) is mounted on the rigid distal phalanx beneath the soft skin for tactile transmission evaluation.}
        \vspace{-1.2em}
    \label{fig:3_skinSensor}
\end{figure}

\subsection{Soft Skin Designs}

The soft skin forms the interface between the rigid finger structure and the manipulated object, and it also affects tactile signal transmission to embedded sensors. 
By developing modular silicone elastomer skins which can be added as a 'sock' to each finger (Fig.~\ref{fig:3_skinSensor}), we can evaluate collective the soft-rigid performance when combining different fingers with varying skins. The skin design parameters are summarized in Table~\ref{table:skins}. The tested materials include Ecoflex and Dragon Skin elastomers (Smooth On Inc.). Each skin was mounted around the rigid phalangeal structure of the assembled tendon and bone system. For tactile transmission evaluation, a force-sensitive resistor (FSR) was attached to the rigid fingertip bone below the skin.

\section{Evaluation Methodology \&  Metrics}

\begin{figure*}[h]
    \centering
    \includegraphics[width=0.99\linewidth]{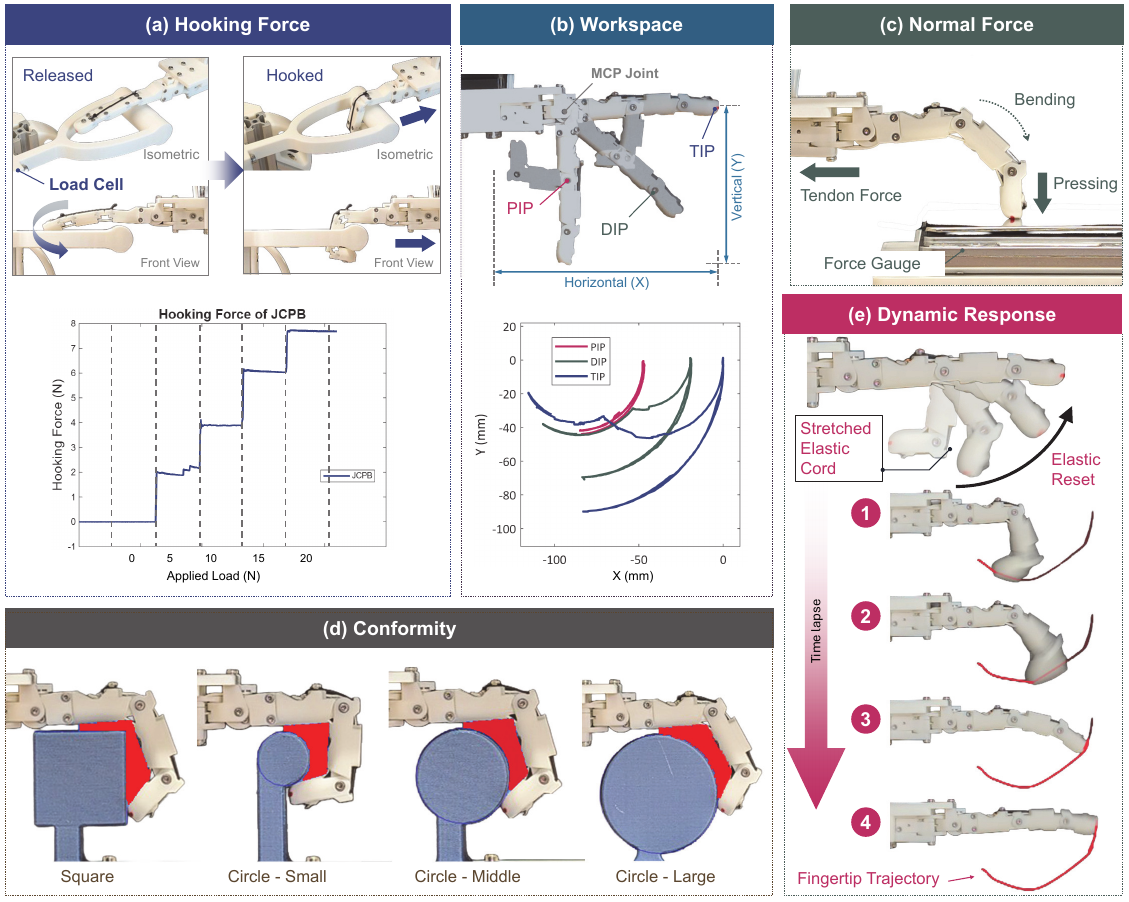}
    \caption{Structural evaluation methods for the finger designs. (a) Hooking force test, showing the released and hooked states, and exemplary time series data under incremental tendon loading. (b) Workspace evaluation, where the trajectories of the PIP, DIP, and fingertip (TIP) markers are tracked in the XY plane to characterise the reachable motion range. (c) Normal force evaluation, in which the finger presses against a force gauge under tendon actuation to assess force transmission performance at various joint configurations. (d) Conformity evaluation with square and circular objects of different sizes, where the red region denotes the gap area between the flexed finger and the object. (e) Dynamic response evaluation, showing the passive extension process after release and the corresponding fingertip trajectory.}
    \label{fig:4_mechTest}
\end{figure*}

At the finger level we first we evaluate the mechanical properties of the skin-less finger design. Secondly, we evaluate the combined skin and bone on contact metrics, namely information propagation to a sensing element and tip contact forces. The selected finger skin combination was then integrated into the dexterous hand and assessed through a number of full-hand benchmarks.  

\subsection{Single Finger Mechanical Metrics}
\label{sec:fingerMetrci}
We now detail the metrics used to assess the mechanical performance of the fingers.

\begin{figure*}[h]
    \centering
    \includegraphics[width=0.9\linewidth]{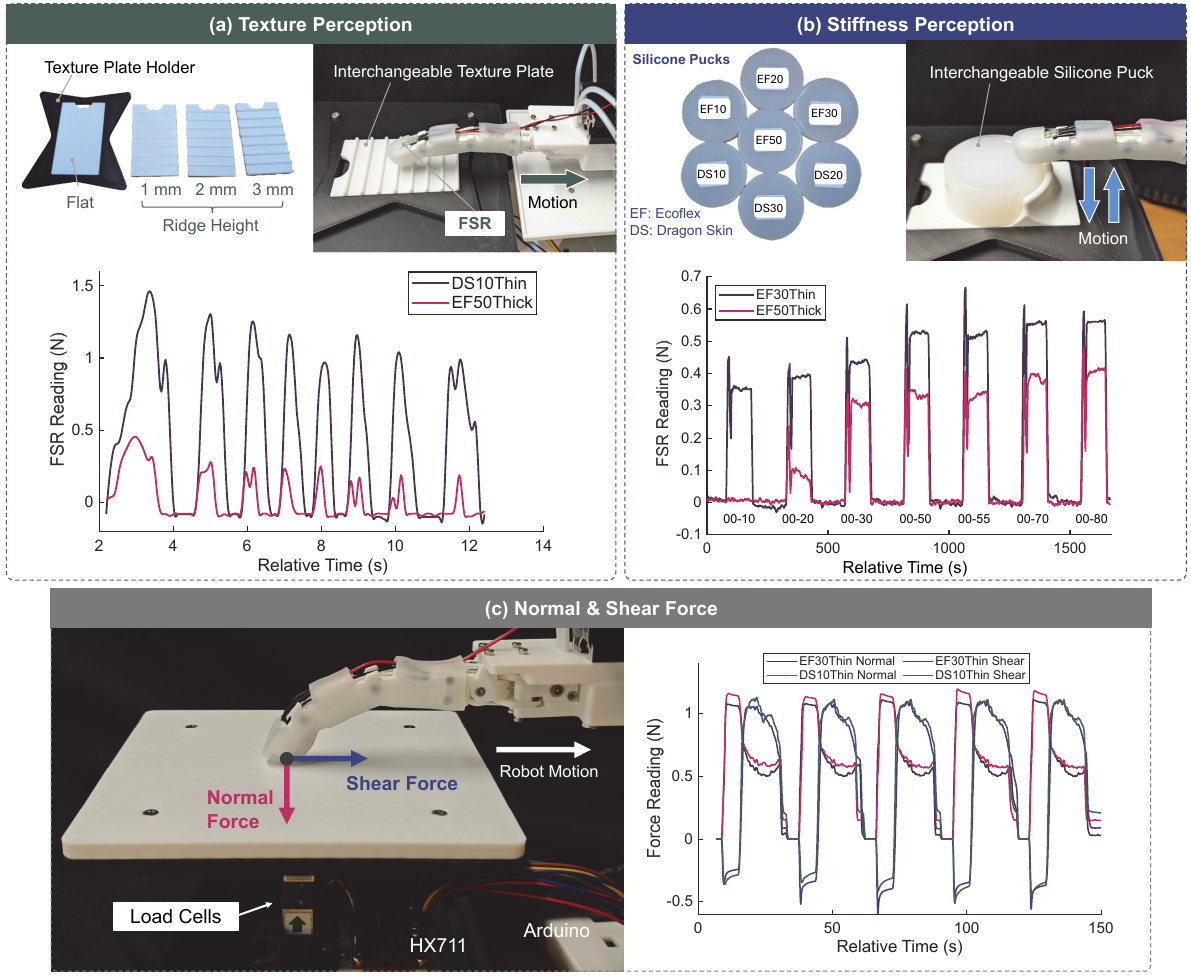}
    \caption{Soft skin evaluation methods. (a) Texture perception test with interchangeable texture plates of different ridge heights and representative FSR readings during one sweep. (b) Stiffness perception test with interchangeable silicone pucks of different stiffnesses and representative FSR readings during seven presses. (c) Normal and shear force test using a load cell platform, with representative force recordings during two experiments.}
    \label{fig:5_skinTest}
\end{figure*}

\paragraph{Hooking Force}
\label{sec:hookForce_metric}
Hooking force capability is important for manipulation tasks such as weight holding and door opening \cite{kashef2020robotic}. We evaluate each finger by applying incremental weights to the driving tendon and measuring the resulting force with a load cell. The setup and a representative force response are shown in Fig. \ref{fig:4_mechTest}a. For the \(j\) th finger design, the maximum stable hooking force before loss of engagement is denoted as \(Q_{\mathrm{HF},j}\).

\paragraph{Workspace}
\label{sec:WS_metric}
The finger must cover a sufficient workspace to support wrapping and grasping across objects of different shapes and sizes. We actuate the MCP joint and the coupled PIP and DIP joints of each design and track the PIP, DIP, and fingertip markers using Kinovea, as shown in Fig. \ref{fig:4_mechTest}b. The workspace score is defined as the mean angular range of the MCP, PIP, and DIP joints. For the \(j\) th finger design, the workspace metric is given as:
\begin{equation}
Q_{\mathrm{WS},j}
=
\frac{1}{3}
\left(
\Delta\theta_{\mathrm{MCP},j}
+
\Delta\theta_{\mathrm{PIP},j}
+
\Delta\theta_{\mathrm{DIP},j}
\right),
\label{eqWorkspaceScoreResult}
\end{equation}
where \(\Delta\theta_{\mathrm{MCP},j}\), \(\Delta\theta_{\mathrm{PIP},j}\), and \(\Delta\theta_{\mathrm{DIP},j}\) are the maximum angular ranges of the three joints.

\paragraph{Normal Force}
The fingertip force generated under a fixed tendon load reflects the combined effect of tendon routing, joint compliance, and transmission losses. To evaluate this property, we built a setup with two orthogonal load cells and a modular loading platform. Each finger is tested under a fixed 20 N tendon load, as shown in Fig. \ref{fig:4_mechTest}c. For the \(j\) th finger design, the maximum normal force measured across the tested heights is denoted as \(Q_{\mathrm{NF},j}\). Repeated trials are averaged.

\paragraph{Conformity}
Human fingers adapt to object shape through the compliance of joints and tendons \cite{cculha2016enhancement}. This conformity is important for robotic hands because it increases contact coverage and improves grasp stability across objects with different geometries. We compare finger conformity by actuating each design from a fixed initial posture under a fixed tendon load around four test objects listed in Table \ref{table:conformity}. The gap area between the flexed finger and the object is measured from captured images. For the j-th finger design, let \(A_{\mathrm{gap},j}\), assumed to be non-zero, denote the mean gap area across objects. The conformity metric is then defined as

\begin{equation}
Q_{\mathrm{CF},j}
=
\frac{1}{A_{\mathrm{gap},j}},
\label{eq:conformityMetric}
\end{equation}
where a larger \(Q_{\mathrm{CF},j}\) indicates better conformity. Fig. \ref{fig:4_mechTest}d shows the measured gap area.

\paragraph{Dynamic Response}
Rapid passive recovery after flexion is a less studied property in the robotic hand community \cite{li2024tactile}, yet it is important for repeated contact and fast flexion-extension transitions. In this study, the motors are identical across all finger designs, so the main difference in passive recovery arises from joint friction, damping, and elastic return. 
We evaluate each design by releasing the finger from full PIP and DIP flexion and tracking the fingertip trajectory, as shown in Fig. \ref{fig:4_mechTest}e. Let \(t_{\mathrm{rec},j}\) denote the recovery time required for the \(j\) th finger to return to a prescribed neighbourhood of its initial position. The dynamic response metric is defined as
\begin{equation}
Q_{\mathrm{DR},j}
=
\frac{1}{t_{\mathrm{rec},j}},
\label{eq:dynamicResponseMetric}
\end{equation}
where, similarly, a larger value indicates faster passive reset.

\subsection{Skin and Sensor Metrics}
\label{sec:skinMetrci}

Whilst the finger 'bone' largely determines the mechanical properties, the localized \textit{contact stability} and capacity to leverage \textit{information from the environment} are determined by the skin and bone.
We thus have a set of metrics that evaluate the bone and skin, these consider the trade-offs between the ability to transmit tactile information for texture and stiffness perception and by their normal and shear force contact stability.

\paragraph{Texture Perception}
Texture-related tactile cues are important for dexterous manipulation and contact interpretation \cite{yousef2011tactile,kaboli2015hand}. We therefore evaluate the extent to which the soft skin preserves or amplifies texture-dependent signals of an embedded force sensor.  The finger is moved over four texture plates with ridge heights of 0, 1, 2, and 3 mm under a fixed tendon load of 20 N, as shown in Fig. \ref{fig:5_skinTest}a. Texture perception is quantified by first computing the mean FSR response for each texture plate and then measuring the average pairwise separation among these texture-specific responses. This metric represents how strongly different surface ridge heights can be distinguished from the sensor output, with larger values indicating greater texture-dependent tactile separability. Let \(\bar{r}^{\mathrm{tex}}_{p,k}\) denote the mean FSR response for the \(p\) th texture plate, where \(p=1,\ldots,P\) and \(P=4\). The texture perception metric is defined as
\begin{equation}
Q_{\mathrm{TP},k}
=
\frac{2}{P(P-1)}
\sum_{1 \leq p < q \leq P}
\left|
\bar{r}^{\mathrm{tex}}_{p,k}
-
\bar{r}^{\mathrm{tex}}_{q,k}
\right| ,
\label{eqTexturePerceptionMetric}
\end{equation}
where a larger \(Q_{\mathrm{TP},k}\) indicates more distinguishable texture dependent sensor responses.

\paragraph{Stiffness Perception}
The capacity to estimate the stiffness of objects being handled is important to modulate grasping in response \cite{drimus2014design}. The skin introduces a trade-off: higher compliance can improve contact conformity and load distribution, but excessive compliance or damping may mechanically filter stiffness-dependent force variations before they reach the FSR. We evaluate whether the soft skin can transmit stiffness-dependent contact information to the embedded sensor. A set of silicone elastomer pucks with identical dimensions and different stiffness values was prepared, as summarized in Table \ref{table:stiffnessPucks}.  Each puck is pressed by a finger skin combination under a fixed tendon load of 20 N (Fig. \ref{fig:5_skinTest}b). 
This metric quantifies how well the finger--skin combination separates pucks of different stiffness in the FSR response space. Let \(\bar{r}^{\mathrm{stiff}}_{\ell,k}\) denote the mean FSR response for the \(\ell\) th puck, where \(\ell=1,\ldots,L\) and \(L=7\). Let \(s_{\ell}\) denote the corresponding Shore hardness. After rescaling both \(\bar{r}^{\mathrm{stiff}}_{\ell,k}\) and \(s_{\ell}\) to \([0,1]\), the normalized response is linearly mapped to the normalized ground truth stiffness as
\begin{equation}
\hat{s}_{\ell,k}
=
a_k \tilde{r}^{\mathrm{stiff}}_{\ell,k}
+
b_k ,
\label{eqStiffnessMapping}
\end{equation}
where \(a_k\) and \(b_k\) are obtained by least squares fitting. The stiffness perception metric is then defined as
\begin{equation}
Q_{\mathrm{SP},k}
=
1
-
\frac{1}{L}
\sum_{\ell=1}^{L}
\left|
\hat{s}_{\ell,k}
-
\tilde{s}_{\ell}
\right| .
\label{eqStiffnessPerceptionMetric}
\end{equation}
A larger \(Q_{\mathrm{SP},k}\) indicates better agreement between sensor response and known stiffness variation.

\paragraph{Normal and Shear Force Capability}
The skin and bone collectively affect contact properties, including tactile sensitivity \cite{yan2024soft}, mechanical compliance, and contact stability, and they also influence the force transmission capability of the finger. We evaluate how different skin designs affect force transmission by first applying a normal load through tendon actuation and then dragging the fingertip tangentially across a flat plate to measure the resulting normal and shear forces, as shown in Fig. \ref{fig:5_skinTest}c. The PIP and DIP joints are actuated under a fixed tendon load of 20 N. Each test is repeated five times. The skin normal force metric and skin shear force metric are defined as:
\begin{equation}
\left\{
\begin{array}{l}
Q_{\mathrm{SNF},k}
=
\frac{1}{R}
\sum_{r=1}^{R}
\max_t F^{\mathrm{N}}_{k,r}(t),
\\[1.0ex]
Q_{\mathrm{SSF},k}
=
\frac{1}{R}
\sum_{r=1}^{R}
\max_t
\left|
F^{\mathrm{S}}_{k,r}(t)
\right| .
\end{array}
\right.
\label{eqSkinForceMetrics}
\end{equation}
where \(R=5\), \(F^{\mathrm{N}}_{k,r}(t)\) is the normal force, and \(F^{\mathrm{S}}_{k,r}(t)\) is the shear force. Larger values indicate greater normal force transmission and stronger shear force generation.

\subsection{Full Hand Metrics}
\label{sec:handMetrci}

\begin{figure}[t]
    \centering
    \includegraphics[width=1\linewidth]{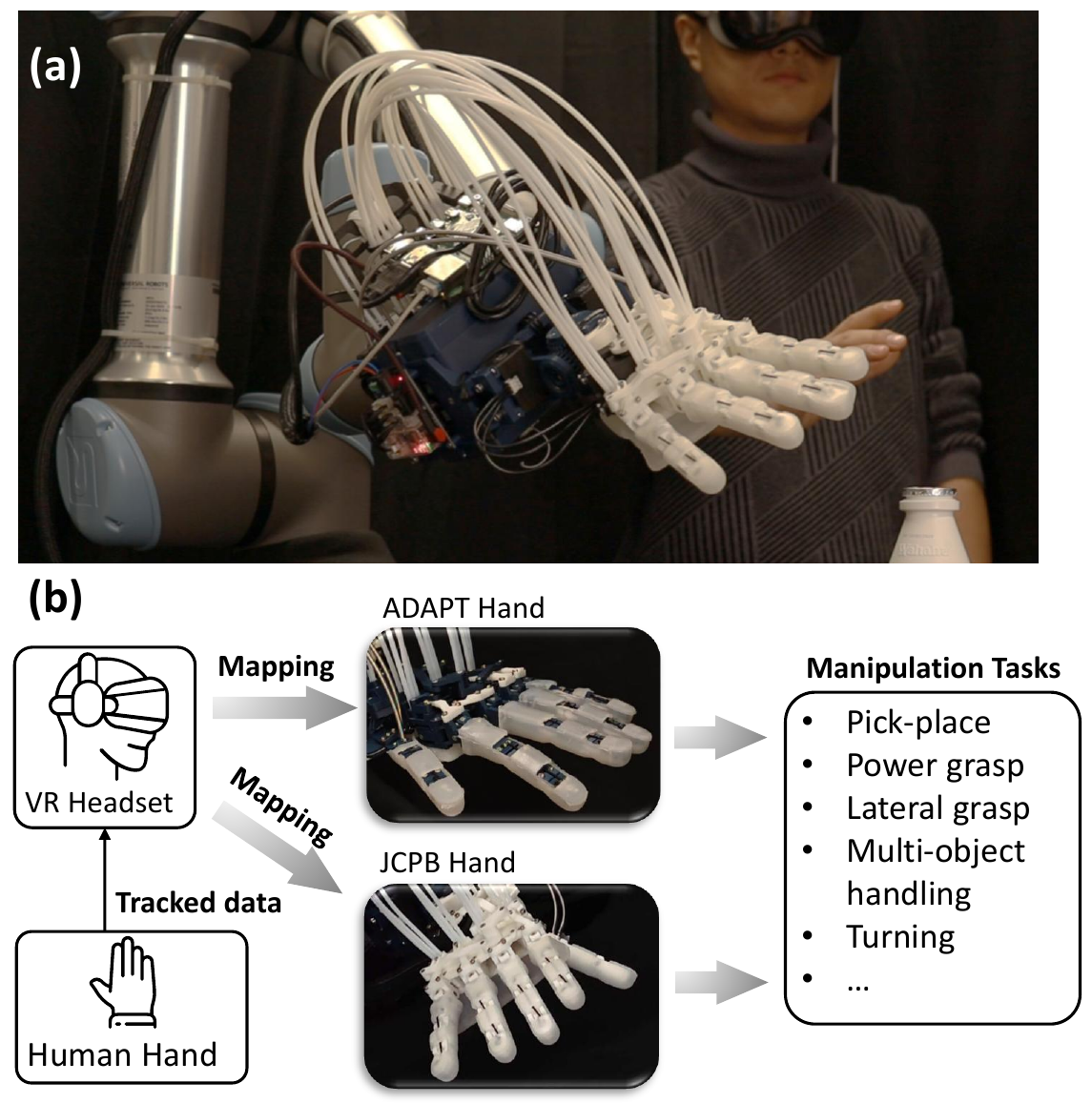}
    \caption{Hand-level evaluation methods. (a) Illustration of the teleoperation setup, in which a human operator remotely controls the robotic system. (b) A VR headset is used to capture human hand motion data, which is then mapped to the corresponding motions of the robotic hand. The robotic hand is evaluated across various manipulation tasks.}
    \label{fig:6_handMethods}
\end{figure}

\begin{figure*}[t]
    \centering
    \includegraphics[width=0.99\linewidth]{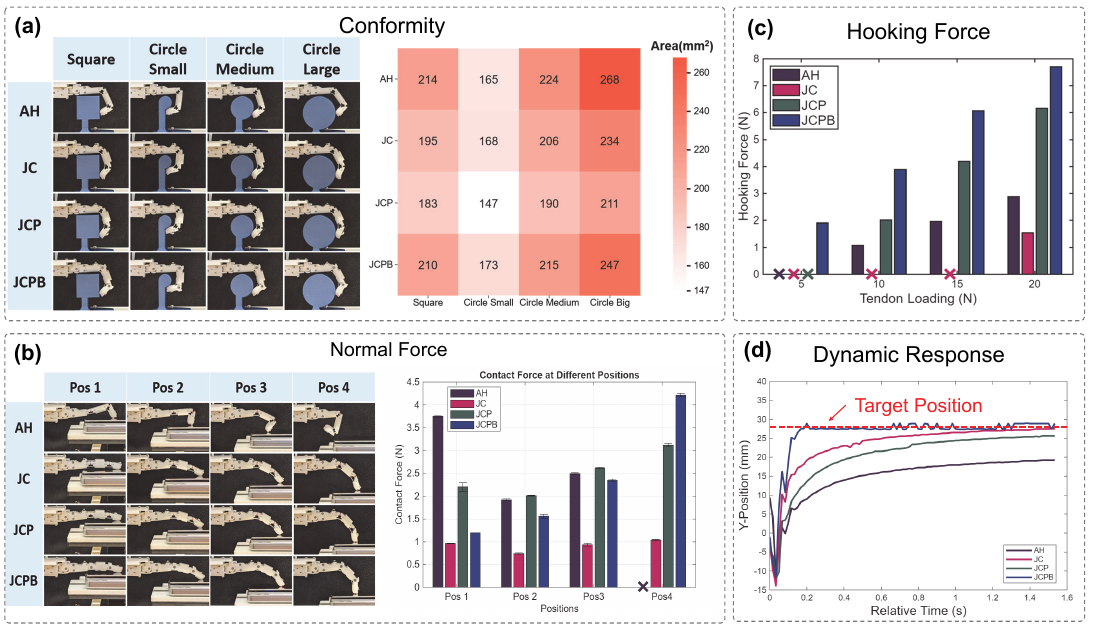}
    \caption{Results of the structural evaluation of the finger designs. (a) Conformity evaluation, with representative snapshots of finger--object wrapping for different object geometries and the corresponding gap-area heatmap. (b) Normal force evaluation, showing representative test configurations at four contact positions and the measured contact force for each finger variant. (c) Hooking force evaluation, where the measured hooking force is a function of tendon loading. (d) Dynamic response evaluation, which depicts the fingertip vertical position during passive return following release from a flexed state; the red dashed line denotes the target position.}
    \label{fig:8_fingerResult}
\end{figure*}

After finger-level evaluation and selection, the chosen finger modules were integrated into a complete robotic hand for hand-level assessment, to ensure that improvements in finger design translate to hand performance. Evaluation was conducted under a teleoperation framework~\cite{junge2025adapt}, in which a VR headset captures the human hand pose and maps it to the MCP and coupled PIP--DIP joint motions of the robotic hand. The hand-level assessment includes both metric-based evaluations, which quantify general manipulation capabilities, and task-oriented evaluations, which examine performance in representative grasping and manipulation tasks.

\subsubsection{Performance Indicator Metrics}
These metrics provide task-independent indicators of full-hand performance and enable comparison with existing state-of-the-art robotic hands. They capture key capabilities relevant to dexterous manipulation, including workspace, force output, grasp stability, and motion responsiveness.

\paragraph{Maximum Lifting Force}
The hand is evaluated in object lifting under three representative grasp configurations, namely thumb index pinching, thumb index middle lifting, and whole hand lifting with the thumb and all four fingers. The lifting force is measured with a load cell. The optimized JCPB hand is benchmarked against the original ADAPT hand.

\paragraph{Maximum Lateral Grasp Strength}
This task evaluates a lateral grasp that relies on frictional contact before lifting. The hand grasps and lifts a bottle with varying water volumes under teleoperation. Up to three successful trials are performed at each water level, and the maximum lifted water mass is used as the metric.

\paragraph{Knob Turning}
This task evaluates rotational manipulation capability \cite{gupta2016learning,morrow2023benchmarking}. The robotic hand is teleoperated to rotate a cylindrical knob (45 mm in diameter). The rotation angle is measured with a rotation encoder (AMS AS5048B), and the total accumulated rotation over five consecutive turns is used for comparison.

\paragraph{Kapandji Scoring}
Thumb opposition is evaluated using a Kapandji scoring test~\cite{kapandji1986clinical}. The robotic thumb is programmed to place its fingertip at multiple target locations across the hand, and higher scores indicate stronger opposition capability.

\subsubsection{Application Driven}

Second, we conduct a set of task-oriented evaluations to assess the hand's teleoperated manipulation performance. These evaluations focus on whether the integrated hand can complete representative manipulation tasks, using task completion time and binary task outcome as the primary measures. Although teleoperation performance can be user-dependent, all trials were performed by the same trained operator after familiarization and repeated under the same protocol, allowing task completion time and success rate to reflect differences in hand performance.

\paragraph{Multi Object Handling}
This task evaluates coordinated manipulation of two objects. The hand first grasps a chewing gum can using the middle, ring, and little fingers together with the hypothenar region, and then pinches a second object with the thumb and index finger. The task is considered successful if both objects are securely held at the end of the trial. The completion time is measured.

\paragraph{Daily Object Grasping}
This task evaluates teleoperated grasping of daily objects with different sizes, shapes, and contact properties. For each object, the hand approaches, grasps, and lifts it from the table. A trial is considered successful if the object is lifted and held without slipping.

\subsection{Evaluation and Selection Process}
\label{sec:scoringSystem}




\begin{figure}[t]
    \centering
    \includegraphics[width=1\linewidth]{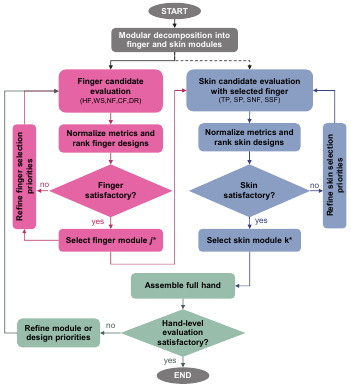}
    \caption{Flowchart of the proposed modular hand design and selection process.}
    \label{fig:flowchart_selection}
\end{figure}

The selection process follows a sequential modular strategy. We first evaluate and rank candidate finger structures, then fix the selected finger structure to evaluate candidate skins, and finally assemble the selected finger--skin configuration into a full robotic hand for hand-level benchmarking.

For finger-level selection, the candidate structures are evaluated using the metrics in Section~\ref{sec:fingerMetrci}. All metrics are normalized across candidates and oriented such that larger values indicate better performance. The finger-level score for design $j$ is
\begin{equation}
\Phi_j
=
\sum_{m \in \mathcal{M}_{\mathrm{F}}}
\beta_m \tilde{Q}^{\mathrm{F}}_{m,j},
\label{eqFingerScoreSimple}
\end{equation}
where $\mathcal{M}_{\mathrm{F}}=\{\mathrm{HF},\mathrm{WS},\mathrm{NF},\mathrm{CF},\mathrm{DR}\}$, $\tilde{Q}^{\mathrm{F}}_{m,j}$ is the normalized metric value, and $\sum_{m \in \mathcal{M}_{\mathrm{F}}}\beta_m=1$. The selected finger structure is
\begin{equation}
j^\star
=
\arg\max_j \Phi_j .
\label{eqSelectedFinger}
\end{equation}

With the finger structure fixed as $j^\star$, candidate skins are evaluated using the metrics in Section~\ref{sec:skinMetrci}, including texture perception, stiffness perception, skin normal force capability, and skin shear force capability. The skin-level score for design $k$ is
\begin{equation}
\Psi_k
=
\sum_{m \in \mathcal{M}_{\mathrm{S}}}
\gamma_m \tilde{Q}^{\mathrm{S}}_{m,k},
\label{eqSkinScoreSimple}
\end{equation}
where $\mathcal{M}_{\mathrm{S}}=\{\mathrm{TP},\mathrm{SP},\mathrm{SNF},\mathrm{SSF}\}$, $\tilde{Q}^{\mathrm{S}}_{m,k}$ is the normalized metric value measured with $j^\star$, and $\sum_{m \in \mathcal{M}_{\mathrm{S}}}\gamma_m=1$. The selected skin is
\begin{equation}
k^\star
=
\arg\max_k \Psi_k .
\label{eqSelectedSkin}
\end{equation}

The selected pair $(j^\star,k^\star)$ defines the final pre-assembly configuration. Its compact overall score is reported as
\begin{equation}
\mathcal{S}_{j^\star,k^\star}
=
\alpha_{\mathrm{F}}\Phi_{j^\star}
+
\alpha_{\mathrm{S}}\Psi_{k^\star},
\label{eq:preAssemblyScore}
\end{equation}
where $\alpha_{\mathrm{F}}+\alpha_{\mathrm{S}}=1$. Unless otherwise specified, equal weights are used. Repeated measurements are reported as mean~$\pm$~standard deviation.

Finally, the selected configuration is integrated into the full robotic hand and evaluated using hand-level indicators and application-driven teleoperation tasks. The optimized hand is benchmarked against the original ADAPT hand configuration. The overall process is summarized in Fig.~\ref{fig:flowchart_selection}.

\section{EXPERIMENTAL RESULTS}
\label{sec4:Results}

\subsection{Finger-Level Raw Results}
\subsubsection{Hooking Force}
Fig.~\ref{fig:8_fingerResult}c shows that hooking force increases with tendon loading for all finger designs. The mean hooking force scores are \(1.46~\mathrm{N}\) for AH, \(0.38~\mathrm{N}\) for JC, \(3.07~\mathrm{N}\) for JCP, and \(4.86~\mathrm{N}\) for JCPB, giving the ranking \(\mathrm{JC}<\mathrm{AH}<\mathrm{JCP}<\mathrm{JCPB}\). JCPB therefore provides the strongest hooking capability.

JC produces measurable hooking force only at the highest tendon load, indicating limited effective force transmission. Adding the integrated pulley increases the output from JC to JCP. Replacing flexible TPU joints with miniature ball bearings further improves force output in JCPB by reducing elastic resistance and viscoelastic losses.

\subsubsection{Workspace}

\begin{figure}[t]
    \centering
    \includegraphics[width=\columnwidth]{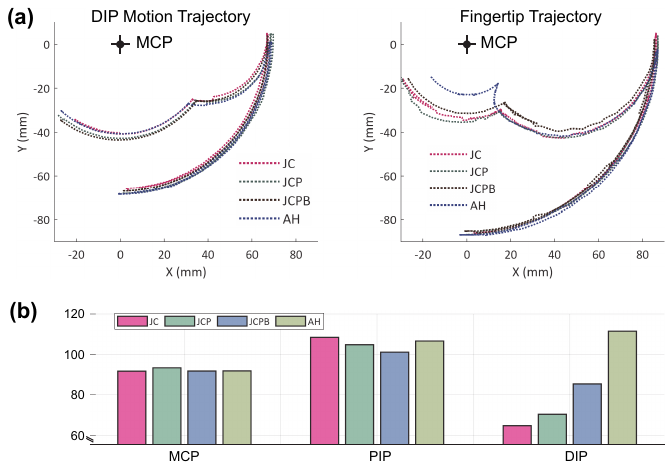}
    \caption{Workspaces of the four finger designs. (a) Trajectories of the tracked DIP and fingertip locations across all four fingers. (b) Comparative analysis of the motion ranges at different joint locations.}
    \label{fig:7_jcpbWS}
\end{figure}

Fig.~\ref{fig:7_jcpbWS} shows the tracked joint and fingertip trajectories and the corresponding angular ranges. The workspace scores are \(103.3^{\circ}\) for AH, \(88.3^{\circ}\) for JC, \(89.5^{\circ}\) for JCP, and \(92.6^{\circ}\) for JCPB. AH achieves the largest overall workspace, mainly due to its larger DIP range. Among the joint-coupled designs, JCPB provides the largest workspace.

The MCP and PIP ranges are similar across the four designs, whereas the main difference arises at the DIP joint. The larger DIP range of AH is enabled by full flexion. In the joint-coupled designs, the passive coupling tendon constrains full DIP flexion and reduces the reachable range. Even so, JCPB preserves a larger workspace than JC and JCP while also providing stronger force output.

\begin{figure*}[h]
    \centering
    \includegraphics[width=0.9\linewidth]{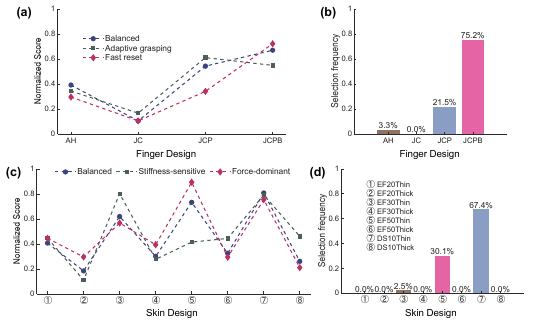}
    \caption{Scores under different weighting conditions. (a) Finger scores under three representative weighting cases. (b) Finger selection frequency from Monte Carlo sampling. (c) Skin scores under three representative weighting cases. (d) Skin selection frequency from Monte Carlo sampling.}
    \label{fig:differentBeta}
\end{figure*}

\subsubsection{Normal Force}

Fig.~\ref{fig:8_fingerResult}b compares normal forces at four contact positions. The mean normal force scores are \(2.04~\mathrm{N}\) for AH, \(0.92~\mathrm{N}\) for JC, \(2.49~\mathrm{N}\) for JCP, and \(2.33~\mathrm{N}\) for JCPB. JCP gives the highest mean value, while JCPB remains close and shows the strongest force output in the most flexed configuration.

The TPU-based fingers show a slight force drop from positions 1 to 2, most clearly in AH, which is likely caused by structural compression and the resulting change in tendon transmission. This effect is not observed in JCPB. JCPB also shows a steady increase from positions 1 to 4 and reaches the highest force at position 4, indicating that the bearing-based joint design is advantageous in highly flexed contact conditions.

\subsubsection{Conformity}

Fig.~\ref{fig:8_fingerResult}a summarizes the conformity results. The mean gap areas are \(217.75\) for AH, \(200.75\) for JC, \(182.75\) for JCP, and \(211.25\) for JCPB. Since a smaller gap area indicates better conformity, JCP shows the best conformity among the four designs.
Although AH has compliant TPU joints and an underactuated structure, it shows the largest gap area. This is mainly caused by over-flexion after contact, which bends the fingertip inward and increases separation along the proximal and middle phalanges. JC and JCP benefit from compliant joints and controlled coupling, while the bearing-based joints in JCPB reduce passive structural compliance.

\subsubsection{Dynamic Response}

Fig.~\ref{fig:8_fingerResult}d shows the fingertip \(Y\) positions during passive extension of the fingers. The dynamic response scores are \(0.18\) for AH, \(0.62\) for JC, \(0.36\) for JCP, and \(5.26\) for JCPB. JCPB clearly outperforms the other designs and provides the fastest passive recovery.

This improvement is mainly attributed to the bearing-based PIP and DIP joints, which reduce frictional loss and viscoelastic resistance during extension. The TPU-based fingers recover more slowly, with AH showing the slowest response.

\subsection{Finger-Level Metrics and Selection}

Based on the scoring model in Section~\ref{sec:scoringSystem}, the raw finger-level measurements were normalized and combined to support pre-assembly selection. Equal weights were first used for all finger metrics.

\begin{table}[h]
\centering
\caption{Normalized finger-level scores.}
\label{tab:normalized_Q_scores_Finger}
\begin{tabular}{lcccc}
\toprule
\textbf{Score} & \textbf{AH} & \textbf{JC} & \textbf{JCP} & \textbf{JCPB} \\
\midrule
\(\widetilde{Q}_{\mathrm{HF}}\) & 0.24 & 0.00 & 0.60 & 1.00 \\
\(\widetilde{Q}_{\mathrm{WS}}\) & 1.00 & 0.00 & 0.08 & 0.30 \\
\(\widetilde{Q}_{\mathrm{NF}}\) & 0.71 & 0.00 & 1.00 & 0.90 \\
\(\widetilde{Q}_{\mathrm{CF}}\) & 0.00 & 0.44 & 1.00 & 0.16 \\
\(\widetilde{Q}_{\mathrm{DR}}\) & 0.00 & 0.09 & 0.03 & 1.00 \\
\addlinespace
\(\widetilde\Phi\) & 1.95 & 0.53 & 2.71 & \textbf{3.35} \\
\bottomrule
\end{tabular}
\end{table}

Table~\ref{tab:normalized_Q_scores_Finger} shows that JCPB obtains the highest finger-level score. Although AH has the largest workspace and JCP gives the best conformity and mean normal force, JCPB provides the strongest overall balance across force transmission, workspace, and dynamic response.

\begin{figure*}[h]
    \centering
    \includegraphics[width=0.9\linewidth]{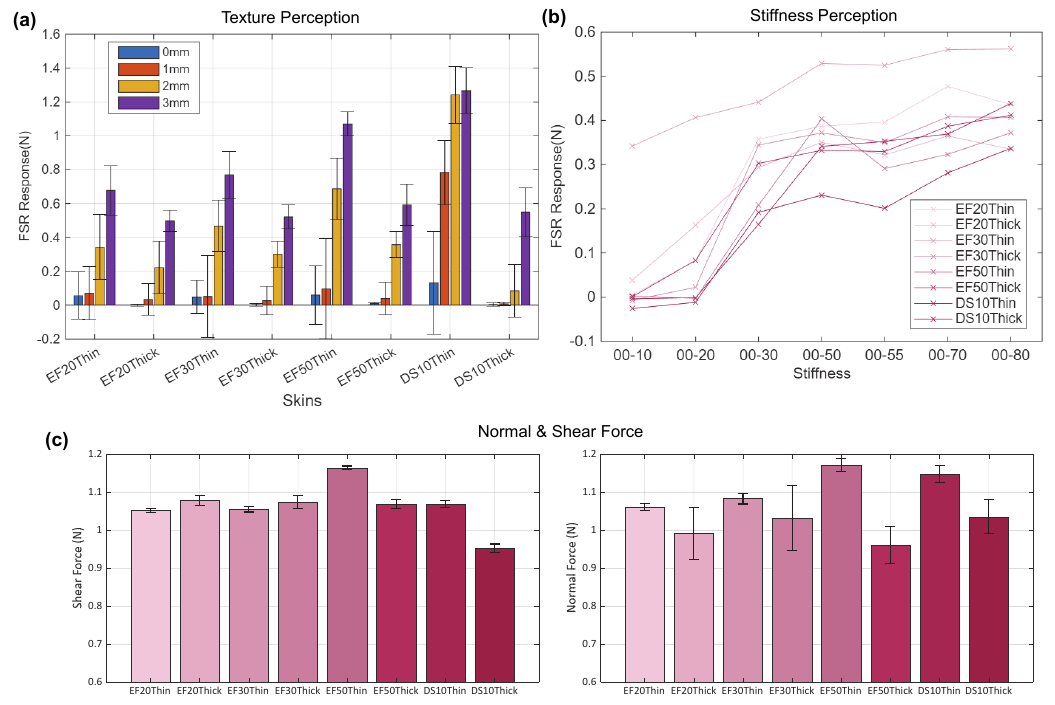}
    \caption{Experimental results in soft skin design evaluations. (a) FSR Responses recorded in the texture perception experiments. (b) FSR Responses recorded in the stiffness perception experiments. (c) Normal and shear forces recorded in the normal and shear force evaluation experiments.}
    \label{fig:9_skinResult}
\end{figure*}

\begin{figure*}[h]
    \centering
    \includegraphics[width=0.9\linewidth]{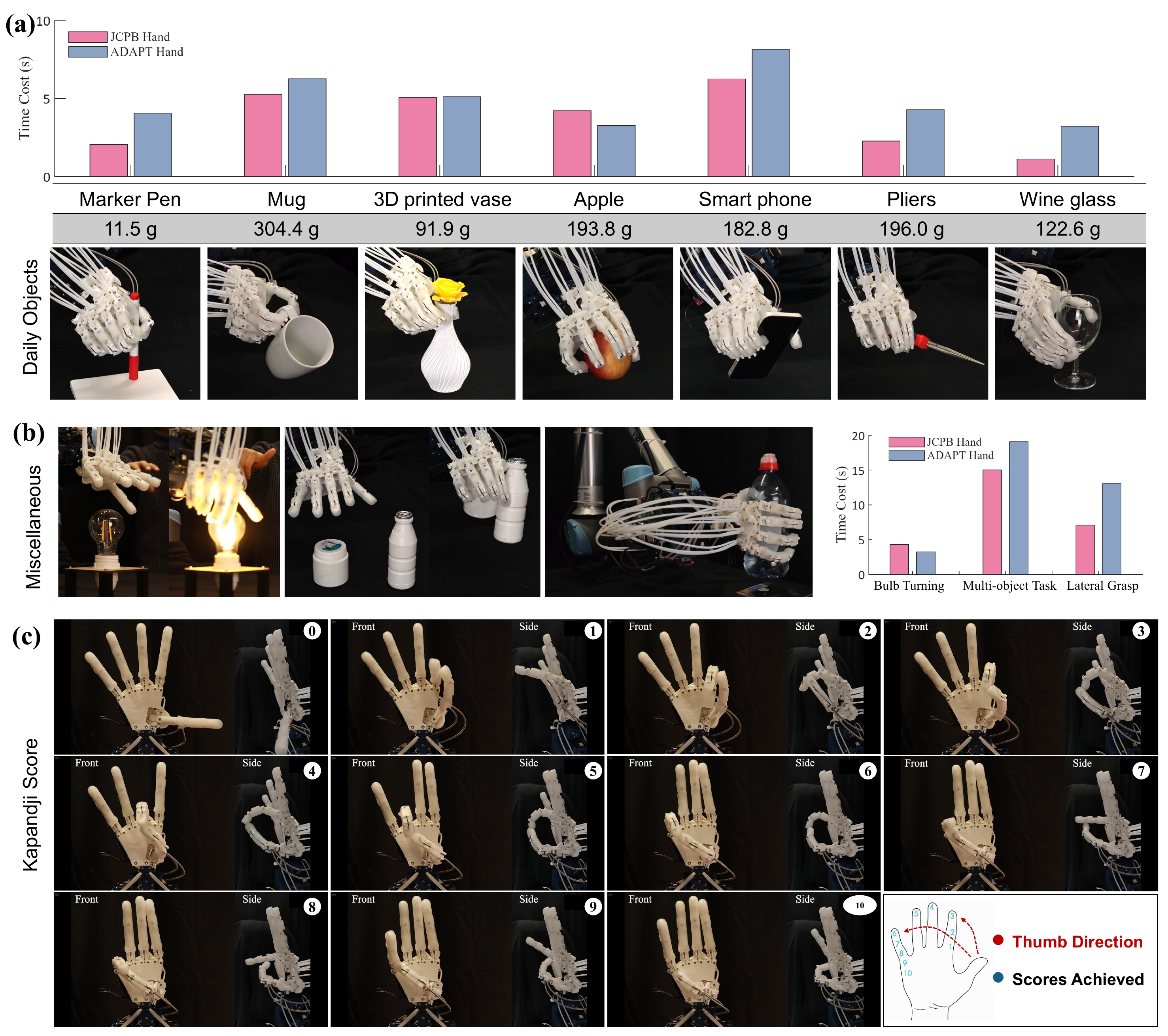}
    \caption{Manipulation performance of the JCPB hand. (a) Pick-and-place of various daily-use objects, (b) representative complex tasks including light bulb turning and multi-object grasping, and (c) evaluation of thumb opposition capability.}
    \label{fig:10_handResult}
\end{figure*}

\begin{figure*}[t]
    \centering
    \includegraphics[width=1\linewidth]{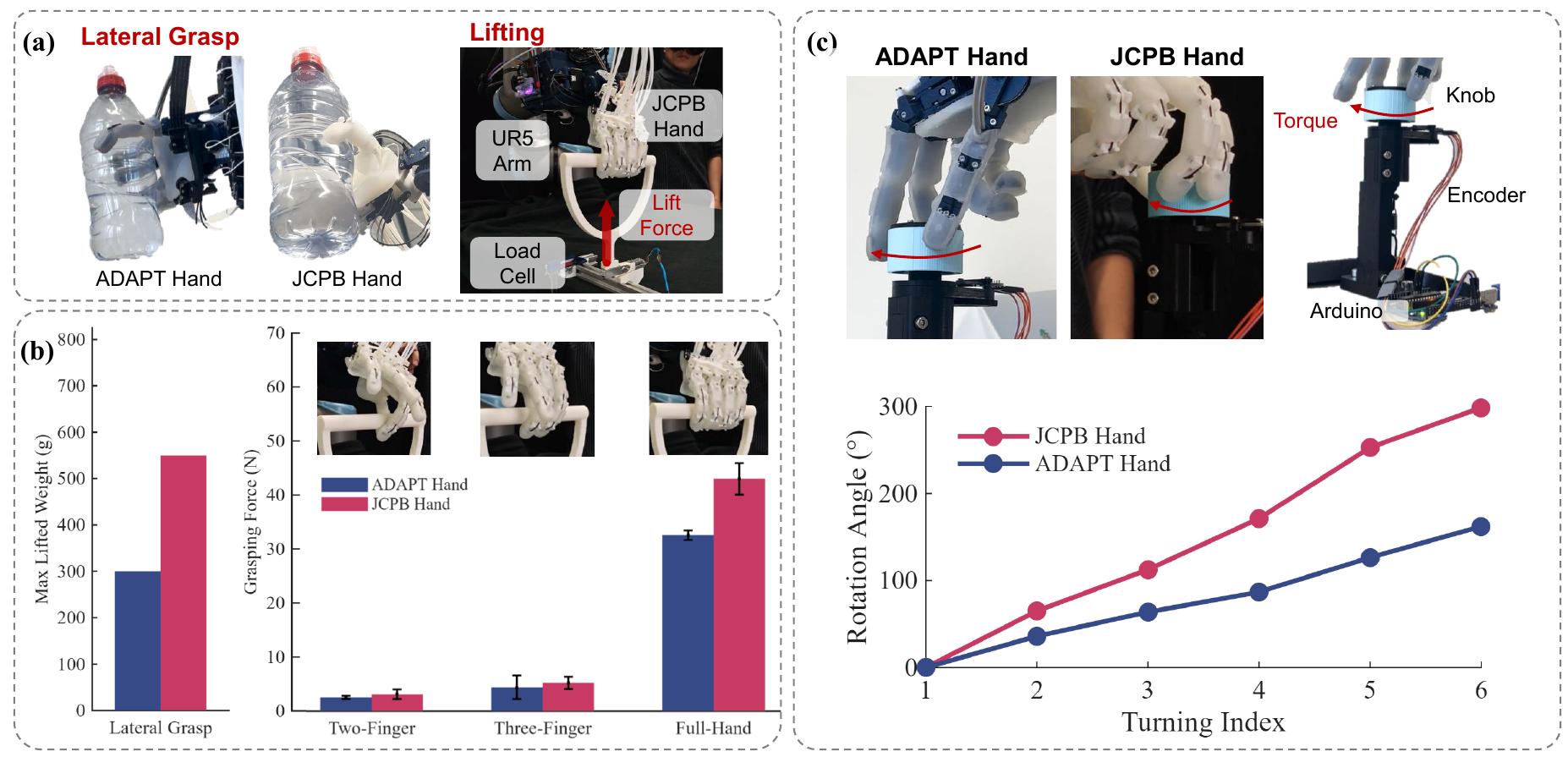}
    \caption{Comparison between the JCPB hand and the ADAPT hand across multiple manipulation metrics. (a) Representative grasping postures, (b) object lifting under different postures, and (c) knob-turning experiment.}
    \label{fig:11_handComp}
\end{figure*}

\subsubsection{Finger Scores under Different Weighting Cases}

In practical manipulation tasks, different performance priorities may require different metric weights. To illustrate this effect, we define three representative combinations of weighting factors \(\beta_m\) in the finger-level scoring model of Equation \ref{eqFingerScoreSimple}. These three cases are summarized in Table \ref{tab:finger_weight_cases}, corresponding to general-purpose balanced design (Case 1), adaptive grasping and wrapping (Case 2), and fast cyclic manipulation (Case 3). Note that these weighting cases are not intended to define a universal optimum, but to illustrate how the proposed scoring model can be adapted to different task priorities.


\begin{table}[h]
\centering
\caption{Metric weights for three representative finger-level evaluation cases.}
\label{tab:finger_weight_cases}
\resizebox{\columnwidth}{!}{%
\begin{tabular}{lccccc}
\toprule
\textbf{Case} & \shortstack{Hook\\Force} &
\shortstack{Workspace} &
\shortstack{Normal\\Force} &
\shortstack{Conformity} &
\shortstack{Dynamic\\Response} \\
\midrule
Balanced & 0.20 & 0.20 & 0.20 & 0.20 & 0.20 \\
Adaptive & 0.15 & 0.20 & 0.15 & 0.35 & 0.15 \\
Fast \& Repeat & 0.10 & 0.20 & 0.10 & 0.15 & 0.45 \\
\bottomrule
\end{tabular}
}
\end{table}

Fig. \ref{fig:differentBeta}a shows the finger scores under three representative weighting cases. Under the balanced case, JCPB achieves the highest score, consistent with the equal-weight result in Table \ref{tab:normalized_Q_scores_Finger}. When workspace and conformity are emphasized for adaptive grasping, JCP becomes the preferred design. When dynamic response is assigned the highest weight for fast cyclic manipulation, JCPB again becomes dominant due to its superior passive recovery speed.

These results show that the preferred finger design depends on task priority. JCPB is favored in general-purpose and fast-reset scenarios, whereas JCP is preferred when adaptive grasping and conformity are emphasized. This comparison further supports the use of a weighted modular scoring model for application-oriented finger selection.

A Monte Carlo sensitivity analysis was further performed by randomly sampling nonnegative metric weights under the sum-to-one constraint. For each sampled weight combination (totals 100,000), the weighted finger scores were computed from the normalized metrics, and the top-ranked design was recorded, as shown in Fig. \ref{fig:differentBeta}b. JCPB was selected most frequently (75.20\%), followed by JCP, whereas AH was selected only rarely and JC was never selected. These results indicate that JCPB is the most robust design across a broad range of task priorities, while JCP remains preferable in a smaller subset of weighting conditions.

\subsubsection{Finger-Bone Selection}

Based on the above raw results, normalized scores, and sensitivity analysis, JCPB was selected as the finger bone for the subsequent skin evaluation and full-hand assembly. This selection prioritizes force transmission and repeatable passive reset, which are important for reliable teleoperated grasping and repeated object interaction. The bearing-based PIP and DIP joints reduce frictional and viscoelastic losses, making JCPB the most balanced finger-bone design for integration with candidate skins.

\subsection{Skin-Level Raw Results}
All skin-level evaluations were performed on the selected JCPB finger structure. This fixed-finger setup isolates the effect of skin design on tactile response and contact-force transmission, while keeping the underlying finger mechanics unchanged.

\subsubsection{Texture Perception}
According to Equation \ref{eqTexturePerceptionMetric}, DS10Thin achieves the highest texture perception score \(\left(Q_{\mathrm{TP}}=0.644\right)\), followed by EF50Thin \(\left(Q_{\mathrm{TP}}=0.603\right)\) and EF30Thin \(\left(Q_{\mathrm{TP}}=0.429\right)\), as shown in Fig. \ref{fig:9_skinResult}a. In general, thin skins outperform their thick counterparts, indicating that increased thickness attenuates texture-dependent tactile signals. For skins with the same thickness, stiffer materials tend to produce larger FSR responses because they deform less and transmit contact variation more directly to the sensor.

\subsubsection{Stiffness Perception}

Fig. \ref{fig:9_skinResult}b shows the stiffness perception results. EF30Thin achieves the highest stiffness perception score \(\left(Q_{\mathrm{SP}}=0.924\right)\), followed by DS10Thin \(\left(Q_{\mathrm{SP}}=0.909\right)\) and DS10Thick \(\left(Q_{\mathrm{SP}}=0.901\right)\). The FSR response generally increases with sample stiffness, indicating that stiffer pucks produce stronger transmitted normal loading at the embedded sensor. Overall, EF30Thin provides the best agreement between sensor response and ground-truth stiffness variation.

\subsubsection{Normal and Shear Force}

Fig. \ref{fig:9_skinResult}c summarizes the normal and shear force measurements. EF50Thin achieves the highest normal force score and the highest shear force score, indicating the strongest force transmission and frictional interaction under the same actuation condition. Thin skins generally produce higher normal force than thick skins, suggesting that increased thickness introduces additional resistance in the joints and reduces transmitted fingertip force. The trend is weaker in shear force, which is also influenced by surface properties and variations during skin installation.

\subsection{Skin-Level Metrics and Selection}

The raw skin-level measurements were normalized and combined using the skin-level scoring model in \ref{sec:scoringSystem}. Equal weights were first used for all skin metrics.

Table \ref{tab:normalized_Q_scores_Skin} shows that DS10Thin achieves the highest total skin score, followed by EF50Thin and EF30Thin. However, EF30Thin was selected for final hand assembly as a more balanced design choice. It provides the highest stiffness-perception score, while maintaining competitive texture transmission and force capability, and offers a softer, more compliant interface for hand-level integration \cite{rus2015design}.

\subsubsection{Skin Rankings under Different Metric Priorities}

\begin{table}[h]
\centering
\caption{Weight coefficients for three representative skin-level evaluation cases.}
\label{tab:skin_weight_cases}
\begin{tabular}{lcccc}
\toprule
\textbf{Case} & \shortstack{Texture\\Perception} &
\shortstack{Stiffness\\Perception} &
\shortstack{Normal\\Force} &
\shortstack{Shear\\Force} \\
\midrule
Balanced & 0.25 & 0.25 & 0.25 & 0.25 \\
Stiffness-sensitive & 0.10 & 0.60 & 0.15 & 0.15 \\
Force-dominant & 0.10 & 0.10 & 0.40 & 0.40 \\
\bottomrule
\end{tabular}
\end{table}


To examine the effect of task preference on skin selection, three representative weighting cases were defined for the skin-level score \(\Psi_k\) in Equation \ref{eqSkinScoreSimple}, as summarized in Table \ref{tab:skin_weight_cases}. Fig. \ref{fig:differentBeta}c shows that DS10Thin is preferred under the balanced case, EF30Thin under the stiffness-sensitive case, and EF50Thin under the force-dominant case. This result indicates that the preferred skin design depends on task priority.   

Similarly, a Monte Carlo sensitivity analysis with randomly sampled \(N=100,000\) non-negative weighting combinations under the sum-to-one constraint was conducted. For each sampled combination, the weighted skin scores were computed from the normalized metrics and the top-ranked design was recorded (Fig. \ref{fig:differentBeta}d). 


\begin{table}[h]
\centering
\caption{Normalized skin-level scores. The best total score is highlighted in bold.}
\label{tab:normalized_Q_scores_Skin}
\begin{tabular}{lccccc}
\toprule
\textbf{Skin} & \(\widetilde{Q}_{\mathrm{TP}}\) & \(\widetilde{Q}_{\mathrm{SP}}\) & \(\widetilde{Q}_{\mathrm{SNF}}\) & \(\widetilde{Q}_{\mathrm{SSF}}\) & \(\widetilde\Psi\) \\
\midrule
EF20Thin  & 0.21 & 0.47 & 0.47 & 0.47 & 1.63 \\
EF20Thick & 0.00 & 0.00 & 0.14 & 0.59 & 0.74 \\
\addlinespace
EF30Thin  & 0.41 & 1.00 & 0.58 & 0.49 & 2.47 \\
EF30Thick & 0.07 & 0.22 & 0.34 & 0.58 & 1.20 \\
\addlinespace
EF50Thin  & 0.89 & 0.04 & 1.00 & 1.00 & 2.93 \\
EF50Thick & 0.18 & 0.57 & 0.00 & 0.55 & 1.29 \\
\addlinespace
DS10Thin  & 1.00 & 0.79 & 0.89 & 0.55 & \textbf{3.23} \\
DS10Thick & 0.01 & 0.68 & 0.35 & 0.00 & 1.05 \\
\bottomrule
\end{tabular}
\end{table}

\subsection{Full-Hand Performance}

Based on the above selection, the optimized finger and skin modules were fabricated and integrated into the robotic hand platform. The resulting JCPB hand was evaluated in real-world manipulation tasks and compared with the previous AH hand.

\subsubsection{Various Manipulation Tasks}

The hand was integrated with a UR5 robot arm and operated under teleoperation~\cite{junge2025adapt}. As shown in Fig.~\ref{fig:10_handResult}, the hand successfully performs pick-and-place of common household objects, as well as more demanding tasks such as bulb turning and multi-object grasping. A Kapandji test further shows that the hand reaches all ten thumb opposition positions, indicating robust thumb opposability and effective palmar workspace.

\subsubsection{Comparison with ADAPT Hand}

Fig.~\ref{fig:11_handComp} compares the optimized JCPB hand with the benchmarking ADAPT hand. In lateral grasping, the maximum sustainable bottle weight reaches 550~g, which is about twice that of the ADAPT hand. In vertical lifting, the JCPB hand produces higher forces in all three grasp configurations and exceeds 40~N in the full-hand grasp. In knob turning, it achieves larger rotation in every trial and exceeds \(300^\circ\) over five turns, again roughly doubling the performance of the ADAPT hand.

These results show that the optimized modular design improves both power grasping and dexterous manipulation at the full-hand level.
\section{Conclusion}        \label{sec5:Conclusion}

This paper presented an iterative modular framework for dexterous robotic hand design and validated it through the development of an anthropomorphic robotic hand. The results show that hand performance can be improved through modular finger design and targeted refinement at the finger level. By further separating the finger into bone and skin modules, the design process becomes more structured, and performance optimization can be carried out through quantitative evaluation of individual modules before full hand integration. In this way, improvements in complex hand-level behavior can be guided by simple and measurable finger-level metrics.

We show that iterative modular design is not only a practical engineering strategy for robotic hand development, but also a scalable methodology for building dexterous systems that are easier to optimize, maintain, and adapt across applications. The proposed quantitative scoring model provides a systematic link from finger and skin benchmarking to module selection and full hand embodiment. Note that the scoring system is built on explicit design metrics, including force transmission, workspace, and texture perception for sensor integration. It can therefore be extended to incorporate additional criteria for future task-specific hand designs. This framework provides a general development methodology for more efficient design iteration in robotic hands and related dexterous robotic systems.

\section*{Acknowledgements} 
This work was supported by Innosuisse, the Swiss Innovation Agency, under Grant 119.806 IP-ENG.
\section*{Appendix} 
\begin{table}[h]
\centering
\caption{Sample parameters in the conformity test}
\label{table:conformity}
\begin{tabular}{llc}
\toprule
\textbf{Variant} & \textbf{Shape} & \textbf{Diameter or Edge Length} \\
\midrule
Square & Square & 35 mm \\
\addlinespace
Circle Small & Circle & 20 mm \\
\addlinespace
Circle Medium & Circle & 35 mm \\
\addlinespace
Circle Large & Circle & 50 mm \\
\bottomrule
\end{tabular}
\end{table}


\begin{table}[h]
\centering
\caption{Material properties and dimensions of the stiffness pucks}
\label{table:stiffnessPucks}
\begin{tabular}{llc}
\toprule
\textbf{Variant} & \textbf{Material} & \textbf{Shore Hardness} \\
\midrule
EF10 & EcoFlex 00 10 & 00 10 \\
\addlinespace
EF20 & EcoFlex 00 20 & 00 20 \\
\addlinespace
EF30 & EcoFlex 00 30 & 00 30 \\
\addlinespace
EF50 & EcoFlex 00 50 & 00 50 \\
\addlinespace
DS10 & DragonSkin 10 Slow & 00 55 \\
\addlinespace
DS20 & DragonSkin 20 & 00 70 \\
\addlinespace
DS30 & DragonSkin 30 & 00 80 \\
\bottomrule
\end{tabular}
\end{table}

\begin{table}[h]
\centering
\caption{Material properties and dimensions of the skin samples.}
\label{table:skins}
\begin{tabular}{llcc}
\toprule
\textbf{Skin Variant} & \textbf{Material} & \textbf{Shore Hardness} & \textbf{Thickness (mm)} \\ 
\midrule
EF20Thin & EcoFlex 00-20 & 00-20 & 3.3 \\
EF20Thick & EcoFlex 00-20 & 00-20 & 4.5 \\
\addlinespace 
EF30Thin & EcoFlex 00-30 & 00-30 & 3.3 \\
EF30Thick & EcoFlex 00-30 & 00-30 & 4.5 \\
\addlinespace
EF50Thin & EcoFlex 00-50 & 00-50 & 3.3 \\
EF50Thick & EcoFlex 00-50 & 00-50 & 4.5 \\
\addlinespace
DS10Thin & DragonSkin 10 & 00-55 & 3.3 \\
DS10Thick & DragonSkin 10 & 00-55 & 4.5 \\
\bottomrule
\end{tabular}
\end{table}

\bibliographystyle{IEEEtran}
\bibliography{referencesOG}

\end{document}